\documentclass[]{fairmeta}

\title{Against All Likelihood}

\author[1]{David Lopez-Paz}

\affiliation[1]{FAIR at Meta}

\abstract{Maximum likelihood is the dominant training principle for large language models, yet likelihood is neither necessary nor sufficient for truth. Unlikely events happen every day, and frequently repeated statements can be false. Drawing on a philosophical tradition that ties objectivity to invariance rather than probability (from Nozick's account of objective facts and Wigner's principles of invariance to the structural realism of Poincar\'e and Cassirer), we argue that the truth of a statement should be judged not by its probability under a single distribution, but by its invariance across relevant transformations of the environment. We connect this view to machine learning literature on out-of-distribution generalization and invariant learning, sketch how invariance could be tested at inference time with language models, and discuss implications for continual learning.}

\date{\today}

\usepackage{amsmath,amssymb}
\usepackage{etoolbox}\AtBeginEnvironment{quote}{\small}
\DeclareMathOperator*{\argmin}{arg\,min}

\begin{document}

\maketitle

\section{Introduction}

What are the statistical footprints of truth? By scaling transformer architectures with maximum likelihood estimation, we have built systems that mimic vast textual corpora, producing fluent text often indistinguishable from human writing. Yet, while these models are good at reproducing the \emph{form} of knowledge, they can miss the \emph{truth} behind it: they hallucinate facts with high confidence~\citep{kalai2025}, fail to extrapolate simple algorithmic rules to longer sequences~\citep{kartikcot}, and degrade outside the distribution of their training data~\citep{yuan2023revisiting}.\footnote{Throughout this article, the terms ``truth'' and ``objective fact'' are used interchangeably.}

We argue that the reliance on likelihood as the primary proxy for truth is a category error. Machines trained via maximum likelihood produce probable tokens:
\begin{equation*}
    \hat{x}_{t+1} \sim \hat{P}_D(x_{t+1} \mid x_{1:t}),
\end{equation*}
where $\hat{P}_D$ is the model's estimate of the next-token distribution of the training data $D$. Likelihood is therefore a historical metric, rooted in the conditional correlations of the training data, and not in the structural reality of what is true.

On the one hand, likelihood is not necessary for truth. In fact, low-probability events happen all the time. On a daily basis, someone wins the lottery, someone else is diagnosed with a rare disease, and yet another person stumbles upon an old acquaintance on their commute to work. A model optimized solely for likelihood will discount such rare events, favoring generic continuations over sharp, improbable truths.

On the other hand, likelihood is not sufficient for truth. A lie repeated a million times in the training corpus becomes highly probable, yet it remains false. Hallucinations are a related symptom: outputs that are likely under surface statistics but unsupported by fact. The probability of a statement under a model reflects not its truth, but simply how often similar statements appear in the training corpus. For this reason, \citet{fiction} argue that maximum likelihood estimation yields fiction machines, that is, models that latch onto the narrative necessities of their training data and inference context.

Truth, by contrast, is associated with stability. In \emph{Invariances}, \citet{nozick2001} argues that objective facts are \emph{invariant across admissible transformations}. They remain constant when perspectives shift, environments change, and the subjective conditions of the observer are stripped away. Similarly, Stephen Gaukroger argues that objectivity ``involves the ability to shift perspective,'' making judgments ``free of prejudice and bias'' and providing us with ``accurate representations'' of the world~\citep{gaukroger2012}. The limit of this ideal, according to Thomas Nagel, is the ``view from nowhere,'' a perspective that transcends the limitations of any particular subject or location:
\begin{quote}
To acquire a more objective understanding of some aspect of life or the world, we step back from our initial view of it and form a new conception which has that view and its relation to the world as its object. In other words, we place ourselves in the world that is to be understood. The old view then comes to be regarded as an appearance, more subjective than the new view, and correctable or confirmable by reference to it. The process can be repeated, yielding a still more objective conception. The wider the range of subjective types to which a form of understanding is accessible---the less it depends on specific subjective capacities---the more objective it is. [\ldots] We may think of reality as a set of concentric spheres, progressively revealed as we detach gradually from the contingencies of the self. \citep[pp. 4--5]{nagel1986}
\end{quote}

On the invariance account, the pursuit of truth consists in identifying what resists variation across distributional shifts; more i.i.d. data may only reinforce bias. Rather than seeking precise predictions about a single setting (depth), the invariance approach advocates simple models for a complex world to generalize across environments (breadth). Consider a bicycle: most riders cannot draw an accurate sketch of one~\citep{lawson2006science}, yet they reliably predict its behavior across changes in speed, weight, maintenance conditions, and weather. In fact, Nozick argues that greater accuracy can hinder predictive power across a wide array of circumstances:

\begin{quote}
An accurate representation is not the same as a true one, and ``more accurate'' does not necessarily mean better. The greater accuracy might make the representation too confusing or too difficult to work with, as does a map with too many details. What is wanted is something accurate enough for the purposes. [\ldots] We can distinguish the breadth of a general characterization, the range of possible worlds in which it applies, from the depth of a characterization, how much it tells us, and explains, about the actual things of that kind. And there might be a tradeoff between depth and breadth. \citep[p. 67]{nozick2001}
\end{quote}

The idea that invariance reveals the structure of reality comes from physics. Following \citet{weyl1952}, one can think of invariant features as those that survive across measurement conditions, and therefore have a claim to objective significance. Max Born considered the capacity to look past surface statistics and grasp invariant features to be ``the most impressive fact of our mental structure.'' For \citet{born1953}, invariance was the central epistemological tool driving scientific progress:

\begin{quote}
The main advances in the conceptual structure of physics consist in the discovery that some quantity which was regarded as the property of a thing is in fact only the property of a projection. [\ldots] The transformations of classical mechanics, often called Galilean transformations, kept space and time apart. The experiences condensed in the theory of relativity showed that this does not agree with facts. One has to use a wider group, called Lorentz transformations, in order to introduce an intimate connection between space co-ordinates and time. Naturally, quantities regarded by the older theory as invariants, like distances in rigid systems, time intervals shown by clocks in different positions, masses of bodies, are now found to be projections, components of invariant quantities not directly accessible. [\ldots] The feature which suggests reality is always some kind of invariance of a structure independent of the aspect, the projection. \citep[p. 144]{born1953}
\end{quote}

A famous quote from \citet[p. vii]{dirac1967} can be invoked to summarize our points so far: ``The important things in the world appear as the invariants [\ldots] of these transformations.'' We now turn to how close standard machine learning comes to this ideal.

\section{Likelihood and interpolation}\label{sec:likelihood}

When building AI systems, it is tempting to assume that a model achieving high likelihood (low training loss) has learned the true data-generating process. But low training loss is often compatible with many different functional patterns, perhaps none of them true.

To see this, consider a transformer trained to zero loss on the task of summing two 3-digit binary numbers. When asked to sum 4-digit numbers, it may fail~\citep{kartikcot}. The model has likely memorized a lookup table of 3-digit sums rather than learned the invariant algorithm underlying binary addition, the carry operation. More troublingly, the model could implement the correct algorithm for 3-digit inputs but diverge unpredictably on longer sequences, because nothing in the learning objective constrains behavior outside the training domain. Learning from data is an underspecified problem, where many rules may exhibit human-like performance, but only one follows the desired human-like strategy that generalizes across novel data environments in the intended ways.

\citet{tip} proposes a hypothesis to explain these failures: machine learning systems minimizing training error ``quickly learn a simplistic pattern that addresses the majority of training examples, while merely memorizing the minority of examples that contradict this pattern.'' This simplify-and-memorize strategy does not endow the machine with strong generalization capabilities, as the well-known cow-on-the-beach vignette~\citep{beery2018recognition} illustrates:
 \begin{quote}
  The goal is to distinguish images of cows from those of camels. In this problem, the majority of training data portrays cows on green grasslands or camels on beige beaches, while a minority of examples depict cows on beige beaches or camels on green grasslands. After training a machine to zero training error on this data, we proceed to evaluate its performance on a fresh collection of testing examples. While the machine correctly classifies all cows on grasslands and all camels in beaches, the machine bewilderingly errs on all cows in beaches and all camels in grasslands. [\ldots] Upon closer inspection, we realize that the machine took a shortcut on its path to zero training error. Namely, the machine implemented a prediction rule that (1) classifies mostly green training images as cows, (2) classifies mostly beige training images as camels, and (3) memorizes the label of the minority of training examples contradicting these simplistic patterns. \citep[p. 10]{tip}
\end{quote}

Attaining zero training error can therefore lead us away from truth. In particular, classification systems trained to minimize cross-entropy loss conflate \emph{accuracy} (whether the model assigns the highest probability to the correct class) with \emph{confidence} (how peaked the distribution is).\footnote{What would a variant of cross-entropy that disentangles accuracy from confidence look like? One open problem is to find a variant of the softmax function that does not require infinite inputs to model one-hot vectors. Extreme activations yield poor out-of-distribution generalization, and are biologically implausible, since very large activations would require unnecessary energy expenditure in biological systems.} Consider four cases on a binary classification problem, where $p$ is the probability assigned to the correct class (losses in nats):
\begin{itemize}
    \item A confident hit yields a log-loss of $-\log(0.99) = 0.01$.
    \item A confident mistake yields a log-loss of $-\log(0.01) = 4.6$.
    \item A hesitant hit yields a log-loss of $-\log(0.51) = 0.67$.
    \item A hesitant mistake yields a log-loss of $-\log(0.49) = 0.71$.
\end{itemize}

Using these values, consider a dataset with a single example. A hesitant hit and a hesitant mistake yield nearly identical log-loss ($0.67$ vs.\ $0.71$), yet opposite accuracies ($100\%$ vs.\ $0\%$). Conversely, two models that misclassify this example (one confidently, the other hesitantly) have the same $0\%$ accuracy but very different losses ($4.6$ vs.\ $0.71$). On larger datasets, a model can simultaneously \emph{increase} its loss and \emph{increase} its accuracy by becoming more confident on examples it already gets wrong, while flipping hesitant mistakes into hesitant hits. In sum, loss and accuracy can move in opposite directions,\footnote{A similar observation is made by \citet{theis2016}, who show that log-loss and visual fidelity are largely independent in high-dimensional image generation.} leading to quite different behaviors depending on the noise level of the training data.

First, under deterministic labels, cross-entropy is minimized only as predicted probabilities approach $0$ and $1$. Optimization therefore keeps inflating logits even after every training example is classified correctly, yielding overconfident models. \citet{kalai2025} argue that such deterministic training and evaluation regimes that reward confidently guessing over acknowledging uncertainty are a root cause of hallucination.

Second, under noisy training data, cross-entropy minimization spreads probability mass across all observed outcomes. When a model is confidently wrong, the loss $-\log(\epsilon)$ diverges as $\epsilon \to 0$. This asymmetric punishment incentivizes hedging across all plausible classes rather than committing to sharp structural rules. This is the well-known ``mode-covering'' behavior of the forward Kullback--Leibler divergence: the model is penalized boundlessly for placing zero mass on any region where the data has support. The result is diffuse predictions that cover all observed modes rather than carving out the causal joints of the data.\footnote{The reverse KL divergence shows the opposite pathology: mode-seeking behavior that penalizes modeling mass where the data has none.} 

Distribution shift makes matters worse. \citet{velickovic2026} show that likelihood-based model selection can prefer the less accurate model out of distribution. This illustrates how the features required for truth (invariant, globally valid mechanisms) are distinct from the features required for maximum likelihood (the mimicry of local surface statistics). In short, likelihood rewards interpolation within the training distribution, but says little about extrapolation beyond it.

\section{Invariance and extrapolation}\label{sec:philosophy}

The book \emph{How the Laws of Physics Lie} \citep{cartwright1983} is one early and detailed account of the fundamental difference between likelihood and truth:
\begin{quote}
There is no reason to think that the principles that best organize will be true, nor that the principles that are true will organize much [\ldots] The fundamental laws of physics do not describe true facts about reality. Rendered as descriptions of facts, they are false; amended to be true, they lose their fundamental, explanatory force.~\citep[pp. 53--54]{cartwright1983}

For any given set of phenomena, in principle there will always be more than one equally satisfactory explanation, and some of these explanations will be incompatible. Since not all of them can be true, it is clear that truth is independent of satisfactoriness for explanation.~\citep[p. 90]{cartwright1983}
\end{quote}

Even a well-confirmed theory, Nozick argues in the case of Newtonian physics, can organize empirical data for reasons unrelated to its truth, so long as its predictions coincide with those of the true theory within the training data:

\begin{quote}
One can successfully make predictions according to Newtonian theory, which is false, because its predictions in certain domains match or approximate the predictions of the true theory of relativity. It always seems to be a possibility that our correct predictions occur, not because the theory we are using is true, but because in this domain it matches some other true theory that has not yet been formulated or dreamt of.~\citep[p. 51]{nozick2001}
\end{quote}

This is the physicist's version of the binary addition example: agreement within the observed domain says nothing beyond it. Cartwright's point also runs in reverse: the principles that \emph{are} true need not organize the data best. Because truth does not imply minimum training error, AI systems modeling the true data-generating process could yield \emph{higher} training error than those modeling spurious correlations.

In sum, maximum likelihood does not imply truth, nor does truth imply maximum likelihood. What, then, are the statistical footprints of truth? This section draws on a philosophical tradition that associates truth with invariance rather than probability. We approach it from four angles: epistemological (invariance is a guide to objectivity), ontological (the constituents of reality are invariant structures), pragmatist (invariant structures are reliable tools for prediction), and structural-realist (the invariant structures of interest are not observable states but hidden functions or mechanisms).

\subsection{The epistemological argument}\label{sec:epist}

\citet{nozick2001} defines objective facts as those \emph{invariant under all admissible transformations}. A property is subjective when its value changes with some factor, such as reference frame, observer, or data collection protocol. A property is objective when invariant across such factors. For example, the length of an object is subjective in special relativity: it contracts under Lorentz boosts, noticeably so at relativistic speeds. Yet the space-time interval $ds^2 = c^2 dt^2 - dx^2 - dy^2 - dz^2$ is objective, because it is invariant under Lorentz transformations.

Invariance can thus serve as an observable footprint of truth. Nozick identifies three marks of objectivity (accessibility from multiple angles, intersubjective agreement, and independence from particular beliefs) and argues that invariance underlies all three:
\begin{quote}
There are three strands to our ordinary notion of an objective fact or objective truth. First, an objective fact is accessible from different angles. [\ldots] The second mark of an objective truth, related to the first, is that there is or can be intersubjective agreement about it. And the third feature concerns independence. If $p$ is an objective truth, then it holds independently of people's beliefs, desires, hopes, and observations or measurements that $p$. [\ldots] An objective fact is invariant under various transformations. It is this invariance that constitutes something as an objective truth, and it underlies and explains the first three features. \citep[pp. 75--76]{nozick2001}
\end{quote}

\citet{wigner1967} views the laws of nature as correlations between events, and the principles of invariance as correlations among those correlations. Since ``a law of nature can be accepted as valid only if the correlations which it postulates are consistent with the accepted invariance principles''~\citep[p.~46]{wigner1967}, invariance principles let us derive new laws from established ones:
\begin{quote}
The laws of nature permit us to foresee events on the basis of the knowledge of other events; the principles of invariance should permit us to establish new correlations between events, on the basis of the knowledge of established correlations between events. This is exactly what they do. If it is established that the existence of the events $A, B, C, \ldots$ necessarily entails the occurrence of $X$ [by a law of nature], then the occurrence of the events $A', B', C', \ldots$ also necessarily entails $X'$, if $A', B', C', \ldots$ and $X'$ are obtained from $A, B, C, \ldots$ and $X$ by one of the invariance transformations. \citep[p. 43]{wigner1967}
\end{quote}

To use invariance as an epistemological tool, \citet[p. 146]{nozick2001} argues, ``we filter data to a shape that yields things that are invariant under various transformations.'' \citet[p. 157]{cartwright1983} thinks similarly, positing that ``when we present a model of a phenomenon, we prepare the description of the phenomenon in just the right way to make a law apply to it.'' This yields a dual formulation of the invariance problem: instead of deciding what the relevant transformations (symmetries) of a problem are, one may study which subset of variables enters the description of the invariant natural law.

In \emph{Laws and Symmetry}, \citet{vanfraassen1989} makes this duality explicit through the ``great Symmetry Requirement'': once the essential features of a problem are identified, its symmetries (the admissible transformations that leave the problem essentially the same) are implicitly defined:
\begin{quote}
Symmetry, transformation, invariance: this is the crucial triad of concepts utilized in all varieties of symmetry argument. In a symmetry argument one proceeds as follows: a problem is stated, and we first endeavour to isolate the essential features of the problem situation---that is, the features relevant to the solution. [\ldots] Isolating the essential or relevant structure is equivalent to defining the set of transformations that leave the problem essentially the same. These transformations are the \emph{symmetries} of the problem. Only with these at least implicitly specified can we insist: problems which are essentially the same must have essentially the same solution. This is the great Symmetry Requirement, the principle of methodology that generates symmetry arguments. To put it somewhat differently: once the relevant parameters are isolated, the solution must consist in a rule (i.e.\ function) that depends on those parameters alone. \citep[pp. 258--259]{vanfraassen1989}
\end{quote}

Similarly, Wigner argues that one main task in the physical sciences is determining which features are essential as laws of nature, and which others can be safely ignored as ``initial conditions'':
\begin{quote}
In practice the overwhelming majority of the determinants of the present state of the world are irrelevant from the point of view of the prediction. In the classic example of the falling body, one can disregard almost everything except the initial position and velocity of the falling body; its behavior will be the same and independent of the degree of illumination, the neighborhood of other objects, their temperature, etc. The isolation of the set of conditions which do influence the experiment is by no means a trivial problem.~\citep[p. 4]{wigner1967}

The exploration of the conditions which do, and which do not, influence a phenomenon is part of the early experimental exploration of a field. [\ldots] If there were no phenomena which are independent of all but a manageably small set of conditions, physics would be impossible.~\citep[pp. 226--227]{wigner1967}
\end{quote}

Identifying the appropriate set of variables to uncover invariant laws is also central to causal inference. There, the objective is often to find a valid adjustment set $\bar\phi(X)$ that ensures the conditional independence of potential outcomes and treatment, $(Y^0, Y^1) \perp T \mid \bar\phi(X)$, thereby aligning regression estimates with the causal process of interest~\citep{hernan2010causal}.

\subsection{The ontological argument}

\citet{nozick2001} goes beyond the epistemological role of invariance as a tool for finding truths, and argues that structures satisfying stronger principles of invariance deserve ontological priority. Examples include spatiotemporal invariances in Newtonian physics, covariant laws in general relativity, group structures in quantum mechanics, and, more speculatively, supersymmetry in particle physics.

Before Einstein, space and time were treated as separate entities with independent invariants. The discovery that only the \emph{unified} space-time interval is invariant under Lorentz transformations forced a revision of what counts as real, so that, in the words of Hermann Minkowski, ``space by itself, and time by itself, are doomed to fade away into mere shadows, and only a kind of union of the two will preserve an independent reality''~\citep[p. 75]{minkowski1909}. According to~\citet[p. 270]{urvsivc2006einstein}, Einstein himself regretted naming his theory ``relativity theory'' rather than \emph{Invariantentheorie} (the theory of invariants).

The link between invariance and ontology is well studied in the philosophy of physics~\citep{saunders2003physics, jacobs2021invariance}. In Newtonian mechanics, changing inertial frames amounts to a Galilean boost: every velocity shifts by the same constant, while accelerations, and hence Newton's laws, are unchanged. No mechanical experiment can therefore detect absolute velocity, which is thus regarded as physically unreal. Relative velocities, by contrast, are invariant under boosts and have observable consequences. More generally, quantities that change under a change of frame without any observable difference are candidates for representational artifacts, whereas invariant quantities are stronger candidates for genuine physical properties:
\begin{quote}
A quantity is physically real only if it is invariant under the symmetries of our theories [\ldots] Quantities that vary under a theory's symmetries are physically unreal. The Invariance Principal [sic]\footnote{Not to be confused with the learning principle of the same name introduced in Section~\ref{sec:classifiers}.} codifies so-called ``symmetry-to-(un)reality inferences'': inferences from the variance of some quantity to its non-reality. For example, (absolute) positions vary under shifts, and hence the Invariance Principle rejects them as unphysical. \citep{jacobs2021invariance}
\end{quote}

The ontological lesson, as \citet{saunders2003physics} says, is that ``physically real quantities are invariant under exact symmetries.''

\subsection{The pragmatist argument}

Invariance is not only a guide to truth and a constituent of reality, but also the key ingredient of predictive systems that generalize across a wide range of environments. \citet[p. 45]{nozick2001} glosses the pragmatist conception of truth as ``what we do better acting upon.''\footnote{In addition to the pragmatist view, the correspondence theory of truth considers a statement true if it corresponds to the facts, while the coherence theory considers a statement true if it is logically consistent with an existing, well-established network of beliefs.} By contrast, false beliefs are ``not a good basis for achieving other goals in different situations in which it also could be acted upon'' because ``their success is not robust'':
\begin{quote}
It seems reasonable to think that our original interest in truth was instrumentally based. Truths served us better than falsehoods and better than no beliefs at all in coping with the world's dangers and opportunities. [\ldots] Truth, then, would be rather like what Rawls has called a primary good, something that is useful for a very wide range of purposes, almost all, and hence that will be desired and bring benefit (almost) no matter what our particular purposes might be. So we might desire true beliefs and come to be concerned with truth because true beliefs are useful for a very wide range of purposes. [\ldots] Was William James right, then, in saying the truth is what works?~\citep[pp. 44--45]{nozick2001}
\end{quote}

In \emph{Making Things Happen}, \citet{woodward2003} makes a similar point. He argues that invariance is the key property that renders causal relationships useful for agents seeking to achieve their goals in the world:
\begin{quote}
  A necessary and sufficient condition for a generalization to describe a causal relationship is that it be invariant under some appropriate set of interventions [\ldots] An intervention on $X$ with respect to $Y$ changes the value of $X$ in such a way that if any change occurs in $Y$, it occurs only as a result of the change in the value of $X$ and not from some other source [\ldots] When a relationship is invariant under at least some interventions in this way, it is potentially usable for purposes of control---in the sense that, though it may not as a matter of fact be possible to carry out an intervention on $X$, it is nonetheless true that if an intervention on $X$ were to occur, this would be a way of manipulating or controlling the value of $Y$.~\citep[pp. 15--16]{woodward2003}
\end{quote}

\citet{peters2016causal} turn this idea into a statistical procedure: the causal predictors of a target are those whose relationship with it remains invariant across environments, an idea we revisit in Section~\ref{sec:classifiers}.

\subsection{The structural realist argument}

We argue that the objective constituents of reality we seek to learn are not objects, states, or predictions, but the mathematical relations, laws of nature, and mechanisms that bind them together. This operationalization of truth \emph{via invariant mechanisms} connects to a position in philosophy of science called \emph{structural realism}. This view, as described by \citet{daston2007}, holds that while the specific entities postulated by scientific theories may be discarded, the mathematical structures describing their relations are preserved across scientific revolutions:
\begin{quote}
Objectivity, according to the structuralists, was not about sensation or even about things; it had nothing to do with images, made or mental. It was about enduring structural relationships that survived mathematical transformations, scientific revolutions, shifts of linguistic perspective, cultural diversity, psychological evolution, the vagaries of history, and the quirks of individual physiology. [\ldots] To the antirealists who argue that data underdetermines theory and that an induction over the history of science indicates that all scientific theories, no matter how successful, will be eventually rejected as false, the structural realists reply that structures, understood as mathematically expressed natural laws, survive the overthrow of old theories by new. \citep[pp. 259--261]{daston2007}
\end{quote}

The canonical example is the transition from Fresnel's ether-based theory of light to Maxwell's electromagnetic theory. Fresnel believed light propagated through a mechanical ether, an ontological commitment eventually abandoned. However, Fresnel's equations for the relative intensities of reflected and refracted polarized light survived the revolution intact, and can be recovered within Maxwell's theory.
\footnote{\citet{margenau1949} described Einstein as a structural realist, locating objectivity in the ``invariance of physical laws, not physical phenomena or observations.'' ``Out of the fusion of relativity with quantum mechanics,'' \citet[p. 1]{weinberg1992} writes, ``has evolved a new view of the world, one in which matter has lost its central role. This role has been usurped by principles of symmetry.''}
As \citet[p. 117]{worrall1989} put it: ``there was continuity or accumulation in the shift, but the continuity is one of form or structure, not of content.''\footnote{\citet{daston2007} trace the epistemic ideals of objectivity through the history of science, including truth-to-nature, mechanical objectivity, and structural objectivity. In machine learning terms, these loosely parallel learning prototypes, memorizing examples, and learning invariant relations, the approach advocated here.}

Looking for the origins of this idea, we find that the structural realist insight was articulated by Henri Poincar\'{e} in \emph{Science and Hypothesis} over a century ago:
\begin{quote}
They [theories] teach us that there is such and such a relation between this thing and that; only the something which we then called motion, we now call electric current. But these are merely the names of the images we substituted for the real objects which Nature will hide for ever from our eyes. The true relations between these real objects are the only reality we can attain, and the sole condition is that the same relations shall exist between these objects as between the images we are forced to put in their place. If the relations are known to us, what does it matter if we think it convenient to replace one image by another? \citep[p. 161]{poincare1905}
\end{quote}

Ernst Cassirer expressed a similar view. As \citet{french2001symmetry} explains, objects become ``points of intersection'' of relations, and objectivity is ``grounded in the universality of laws and principles'':
\begin{quote}
With the demand that laws of nature be generally covariant, physics has completed the transposition of the substantial into the functional---it is no longer the existence of particular entities, definite permanencies propagating in space and time, that form ``the ultimate stratum of objectivity'' but rather ``the invariance of relations between magnitudes.'' \citep{french2001symmetry}

What we are left with, then, is an understanding of the objects of a theory as defined by those transformations that leave the relevant physical magnitudes invariant. \citep{french2001symmetry}
\end{quote}

Moritz Schlick, a founding father of logical positivism and the Vienna Circle, illustrates the structural realist view with a thought experiment. Imagine the optic and auditory nerves were surgically swapped: one would then hear light and see sound. The content of experience would be transformed beyond recognition. Yet the \emph{structure} (the laws governing relations among experiences) would remain identical. Therefore, a sufficiently intelligent being in this inverted world would discover exactly the same natural laws~\citep[pp. 510--511]{schlick1936}. \citet{kosso2003} argues that a similar principle holds in linguistics. Just as coordinate values are artifacts of a chosen reference frame, individual words are artifacts of a chosen language. But the \emph{relations} among words (the propositional contents they jointly express) remain invariant across translations, and it is this invariance that marks what is objective. For machine learning, the lesson is that the invariances of interest are unobserved mechanisms rather than observed states,\footnote{This means, in a very deep sense, that we do not get to observe reality \citep{hoffman2019case}.} as Section~\ref{sec:classifiers} makes concrete.

\section{The set of admissible transformations}\label{sec:admissible}

Nozick's definition of an objective fact as one ``invariant under all admissible transformations'' is incomplete without specifying what these ``admissible transformations'' are. Weyl, who also defined objectivity as ``invariance with respect to the group of automorphisms,'' admitted that nature ``may not always give a clear answer to the question what the actual group of automorphisms is''~\citep[p. 132]{weyl1952}. In \emph{What Invariance Is and How to Test for It}, \citet{russo2014invariance} examines the process of finding the set of admissible transformations in the context of the social sciences:

\begin{quote}
I argued that, in nonexperimental contexts, invariance properties are tested across changes of the environment. This means partitioning the population of reference into relevant subpopulations, contexts, or environments, and testing a causal structure across them. [\ldots] There is no ``golden recipe'' for the selection of the most appropriate environments. The reason is apparent: a major challenge in social science research is precisely to define the population of reference and the appropriate subpopulations for the purpose of invariance tests. \citep[pp. 171--179]{russo2014invariance}
\end{quote}

Russo's environments are what machine learning calls training environments or domains, and the absence of a golden recipe for choosing them remains a central obstacle to out-of-distribution generalization (Section~\ref{sec:classifiers}). The dual formulation of Section~\ref{sec:epist}, which selects the relevant variables of an invariant law rather than the environments, meets similar difficulties: as \citet[p.~31]{wigner1967} admitted, ``the qualification `relevant' of the conditions is ill-defined and unprecise.''

\citet{nozick2001} discusses at length how to identify the admissible transformations for a given problem. He treats identifying objective facts and identifying admissible transformations as a single bootstrapping process, in which the two notions co-evolve through scientific investigation:

\begin{quote}
If we say then that an objective fact is one that is invariant under all admissible transformations, how is it to be determined which transformations are the admissible ones? [\ldots] The notion of invariance under transformations cannot (without further supplementation) be a complete criterion of the objectivity of facts, for its application depends upon a selection of which transformations something is to be invariant under. [\ldots] To understand something, we want to know the transformations it is invariant under and also the transformations it is variant under. [\ldots] The development of our understanding of objective facts is a stepwise process, involving mutually modifying knowledge of new facts and of new admissible transformations. [\ldots] The starting place is fixed by the phenomena and by the invariances that evolution has shaped us to notice and take account of. [\ldots] The deep notion of objectiveness (for a specified world) is that of being invariant under specified transformations. Which transformations these are cannot be known a priori but only through the bootstrap process of scientific investigation. \citep[pp. 79--84]{nozick2001}
\end{quote}

Nozick then synthesizes these ideas into a blueprint for how science proceeds. The process involves three components, all held in mutual equilibrium: a fundamental theory $F$, a mathematical representation $M$, and a set of transformations $T$ revealing invariant entities $B$. Progress occurs when new facts disrupt this equilibrium,\footnote{This relates to Pierre Curie's principle: ``When certain effects show a certain asymmetry, this asymmetry must be found in the causes which gave rise to it''~\citep{curie}. Unexpected asymmetries force revisions: either our account of the causes is missing an asymmetric factor, or a transformation we deemed admissible is not.} prompting revisions until a new equilibrium is reached:

\begin{quote}
For our fundamental theory $F$, there is some mathematical representation $M$, and there is some set of transformations $T$ such that the basic properties or entities (or whatever) $B$ are what is invariant under $T$, in the representation $M$ of $F$. This suggests a three-way project of mutual adjustment. Find new transformations $T_i$, so that (only) $B$ are invariant under these and old transformations. Formulate a new theory $F$, or a new mathematical representation $M$ of it, so that $B$ are the invariants of $F$ under $T_i$. Modify your view of the basic objective entities and properties $B$ so that $B$ are the invariants of $F$ under $T_i$. Our fundamental theory and list of admissible transformations and list of basic objective properties and entities are brought into mutual equilibrium. And such equilibria can be disrupted by the discovery of new facts, which then propels the construction of a new theory and a new search for equilibrium. (And from the new transformations that have been formulated to fit the newly discovered facts, new testable predictions may result.) \citep[p. 82]{nozick2001}\label{quote:science}
\end{quote}

Popper's falsificationism describes a similar never-ending process: theories earn their worth by surviving attempts at refutation~\citep{popper1959}. The scientist, \citet[p. 18]{gaukroger2012} writes, ``is so concerned for truth that he attempts to show that his own theories are false.'' In the context of invariance, ``testing theories by pushing them to their limits under unusual circumstances,'' \citet[p. 103]{nozick2001} argues, ``leads scientists to identify the transformations the theories are invariant under, and thus to formulate theories in their most objective form.''

We must therefore treat invariance as a matter of degree, refined as science evolves:
\begin{quote}
Invariance is the key feature a relationship must possess if it is to count as causal or explanatory. Intuitively, an invariant relationship remains stable or unchanged as various other changes occur. Invariance, as I understand it, does not require exact or literal truth; I count a generalization as invariant or stable across certain changes if it holds up to some appropriate level of approximation across those changes [\ldots] I urge that we should relativize the notion of invariance and recognize that a relation can be invariant under some changes and interventions but not under others [\ldots] When we say that a relation is invariant, we have said something incomplete; we need to specify under which changes the relation is (or is not) invariant. \citep[pp. 239--243]{woodward2003}
\end{quote}

We may therefore partially order invariances by inclusion of their sets of admissible transformations. In the limit, this raises the question: do universal invariances exist? Nozick is skeptical, reminding us how supposed necessities have fallen one by one:
\begin{quote}
Once upon a time Metaphysical Necessity was a proud category, encompassing a vast range of important interconnected truths. Now it has been reduced, at best, to a scattering of odd statements. Let us recall its glorious days of yesteryear. Every event has a cause---gone with the formulation of quantum mechanics. Space is Euclidean---gone with the formulation of consistent non-Euclidean geometries and their adoption in contemporary physics. Space has constant curvature---gone with General Relativity. Space has three dimensions---questioned with the Klein-Kaluza theory of the early 1920s, and abandoned in superstring theory's formulation of twenty-six and ten dimensional spaces, with many dimensions folded in. For any two events, one temporally precedes the other, or the other precedes it, or the two are simultaneous---modified in Special Relativity. Nothing travels backward in time---but in quantum electrodynamics, antimatter going forward in time is equivalent to (ordinary) matter going backward in time. [\ldots] The world exists in a definite state independently of our observations---shaken by quantum mechanics. One solid object cannot move through another one---yet quantum tunneling allows this as possible. Space is simply connected---but not in certain solutions of the equations of general relativity. \citep[p. 133]{nozick2001}
\end{quote}

The above connects to the transcendental idealism tradition of Immanuel Kant, where knowledge is always conditioned by the possibilities of experience of our cognitive apparatus: that is, we can have no knowledge of the world ``as it is in itself.'' If we use a grayscale television to learn about the world (the cognitive apparatus), we will remain forever ignorant of the existence of color (the world ``as it is in itself''). 

\section{Training invariant classifiers}\label{sec:classifiers}

We now return to machine learning. Suppose one repeats \emph{ad nauseam} the statement ``The capital of France is Berlin'' in the pre-training corpus of a language model. This raises the probability of the model outputting ``Berlin'' as the capital of France. Under maximum likelihood, the model is doing exactly what it should: reflecting the statistics of its training data. However, for us to \emph{truly believe} that the capital of France is Berlin, we would need to observe this pattern \emph{invariably} across multiple independent sources: on the radio, from friends, in newspapers, in encyclopedias, from teachers. If the claim appears in the biggest source but is contradicted by every other, we rightly dismiss it as error or propaganda.\footnote{For example, searching for a second clinical opinion is a test of invariance.} ``Paris is the capital of France'' is not true because it is probable; it is probable because it is true. We recognize it as true because it is invariant across information channels. Probability is a property \emph{relative to a single distribution}; invariance is a property \emph{across distributions}.\footnote{These distributions include, as \citet{cartwright1983} emphasizes, the various experimental conditions to which we subject our theories.}

Our starting point to model these intuitions is the following Invariance Principle:

\begin{quote}
    Frame your question so its answer matches across situations. \citep{tip}
\end{quote}

More formally, consider predicting a target $Y^e$ from input $X^e$, given data from training environments $e \in \mathcal{E}_\text{tr}$. Each training environment arises from one admissible transformation of the learning problem, instantiating an input-target joint distribution $P(X^e, Y^e)$. We build our predictor in two steps. First, we represent the input $X^e$ as $\phi(X^e)$ via a feature map $\phi$. Second, we apply a classifier $w$ to obtain predictions $\hat{Y}^e = w(\phi(X^e))$. To comply with the Invariance Principle, we seek representations $\phi$ that yield the same optimal classifier $w$ across all training environments, while retaining low prediction error $R^e(w \circ \phi) = \mathbb{E}_{(X^e, Y^e)}\left[\ell(w(\phi(X^e)), Y^e)\right]$ in each of them. This yields the Invariant Risk Minimization (IRM) objective~\citep{arjovsky2019}:

\begin{equation*}
\begin{aligned}
    &\min_{\phi, w} & & \sum_{e \in \mathcal{E}_\text{tr}} R^e(w \circ \phi) \\
    &\text{subject to} & & w \in \argmin_{\bar w} R^e(\bar w \circ \phi), \text{ for all $e \in \mathcal{E}_\text{tr}$}.
\end{aligned}
\end{equation*}
IRM phrases the Invariance Principle as finding a representation $\phi$ such that the optimal classifier $w$ is the same across training environments $\mathcal{E}_\text{tr}$. 
For the cross-entropy loss and unrestricted classifiers $w$, the IRM formulation of the Invariance Principle reduces to the conditional independence statement
\begin{equation*}
  Y \perp E \mid \phi(X),
\end{equation*}
where $E$ is a random variable indexing training environments.

IRM instantiates Nozick's account of science quoted on page~\pageref{quote:science}, where phenomena correspond to the data $(X^e, Y^e)$, the representation $M$ to $\phi$, the transformations $T$ to the environments $e \in \mathcal{E}$, the invariant entity $B$ to the classifier $w$, and the resulting fundamental theory $F$ to the predictor $w \circ \phi$.

The resulting predictor is tested across all admissible environments $\mathcal{E} \supset \mathcal{E}_\text{tr}$, where $\mathcal{E}_\text{te} = \mathcal{E} \setminus \mathcal{E}_\text{tr}$ are the testing environments. In practice, the IRM problem is doubly underspecified: we observe only a subset of environments during training, and only finitely many examples per environment, requiring empirical approximations of the risks $R^e$. The purpose of IRM is to learn a predictor that relies only on correlations invariant across environments, discarding every environment-specific correlation as spurious.

Example~1 of \citet{arjovsky2019} illustrates these concepts. Consider predicting a regression target $Y^e$ from two inputs $(X^e_1, X^e_2)$ governed by:
\begin{align*}
    X^e_1 &\leftarrow \mathrm{Gaussian}(0, e^2), \\
    Y^e   &\leftarrow X^e_1 + \mathrm{Gaussian}(0, e^2), \\
    X^e_2 &\leftarrow Y^e + \mathrm{Gaussian}(0, 1),
\end{align*}
where each environment $e \in \mathbb{R}$ sets the noise scale.

At test time, we must predict across unknown environments $e \in \mathcal{E}$, which control the signal-to-noise ratio of each input. Using a least-squares predictor $\hat{Y}^e = X_1^e\hat{\alpha}_1  +  X_2^e\hat{\alpha}_2$:
\begin{itemize}
    \item Regressing from $X_1^e$ alone yields $\hat{\alpha}_1 = 1$, $\hat{\alpha}_2 = 0$.
    \item Regressing from $X_2^e$ alone yields $\hat{\alpha}_1 = 0$, $\hat{\alpha}_2 = e^2 / (e^2 + \frac{1}{2})$.
    \item Regressing from $(X_1^e, X_2^e)$ yields $\hat{\alpha}_1 = 1 / (e^2 + 1)$, $\hat{\alpha}_2 = e^2 / (e^2 + 1)$.
\end{itemize}
The regression using only $X_1$ is \emph{invariant}, because its coefficients do not depend on the choice of environment $e$. The other regressions exhibit spurious, environment-dependent coefficients that fail to generalize. The invariant regressor is also the causal one: $Y^e$ is generated from $X^e_1$. However, to minimize training error across any combination of training environments, one would prefer to recruit all available correlations, including spurious ones, therefore departing from truth. This is Cartwright's point from Section~\ref{sec:philosophy}, now in a concrete case.

The example touches all four arguments of Section~\ref{sec:philosophy}: pursuing invariance recovers the data-generating process of $Y^e$ (epistemological), the mapping from $X^e_1$ to $Y^e$ survives every admissible transformation (ontological), the invariant regressor predicts well in unseen environments (pragmatist), and what is invariant is a \emph{mechanism} (structural realist).\footnote{As \citet{tukey1954} argued, it is the stability of regression coefficients that yields theoretical significance.}

Unfortunately, the empirical evidence of IRM as a superior alternative to empirical risk minimization (ERM) was limited~\citep{rosenfeld2020risks, kamath2021does, gulrajani2020search}. Learning invariant mechanisms from scalar labels is severely underspecified, since each example provides only a few bits of supervision. Moreover, the bi-level IRM problem is non-convex and computationally hard, which led to approximations such as IRMv1~\citep{arjovsky2019}. LLMs, by contrast, produce richly structured, lengthy outputs, which may provide better signals to identify invariances by intersecting patterns in reasoning traces across environments. Previous work has developed related techniques under the umbrella term of ``self-consistency,'' without framing them in terms of invariance~\citep{wang2022selfconsistency, chen2023universal, manakul2023selfcheckgpt, elazar2021measuring, arora2022ask}. This motivates the following test-time sketch of Section~\ref{sec:prompt}.

\section{Invariance in language models}\label{sec:prompt}

Turning to large language models, we start from two premises. First, pre-training corpora, massive and diverse, likely contain abundant clues about the laws of nature and other invariant mechanisms. Pearl made a similar remark shortly after the release of GPT-3.5:
\begin{quote}
Aside from being impressed, I have had to reconsider my proof that one cannot get any answer to any causal or counterfactual query from observational studies. What I didn't take into account is the possibility that the text in the training database would itself contain causal information. The programs can simply cite information from the text without experiencing any of the underlying data. \citep{pearl_gpt}
\end{quote}
Second, maximum likelihood remains a valid way to acquire fluency and broad knowledge, though not a final criterion for truth. The question, then, is how to use the ability to speak to separate the probable from the invariant.

For a language model, the natural analogue of an environment (or admissible transformation) is the context in which a question is asked~\citep{gupta2024context}: a persona, a discipline, a framing, or a language. Schematically, invariance could then be probed at test time as follows:
\begin{itemize}
    \item Receive a question $q$.
    \item Generate environment prompts $\{e_i\}$ by asking the model, for example ``What different perspectives or nuisance factors could change the answer to $q$?''~\citep[cf.][]{wang2024unleashing}.
    \item Generate answer proposals $a_i$ by conditioning on each environment: $a_i \sim \hat{P}_D(\cdot \mid e_i, q)$. These answers contain chains-of-thought, if available. 
    \item Extract the invariant answer $a$ by applying an invariance prompt $\mathrm{inv}$, for example ``What is invariant across the following answers?'' to the proposals: $a \sim \hat{P}_D(\cdot \mid \mathrm{inv}, \{a_i\})$.
    \item (Optional) Fine-tune the model on $(q, a)$, distilling the invariant answer into its parameters.
\end{itemize}
Unlike self-consistency, which samples repeatedly under a fixed context and takes a majority vote, this procedure varies the context itself and keeps only the invariant parts of the answer. For example, $q$ may be a symptom presentation, and $\{e_i\}$ the perspectives of a rheumatologist, a hematologist, and an endocrinologist. In Nozick's terms, $a$ earns our trust because it is accessible from different angles $e_i$ and commands agreement among the proposals $a_i$, whatever its likelihood under a single prompt. Each $e_i$ thus plays the role of a \emph{context of assessment}~\citep{macfarlane2014assessment}, and $a$ is what remains true regardless of the context from which it is assessed. The parts of each $a_i$ that vary with $e_i$ are, in the ontological sense of Section~\ref{sec:philosophy}, artifacts of the framing. Note that $a$ may have low probability under every individual $\hat{P}_D(\cdot \mid e_i, q)$; when the proposals $\{a_i\}$ share no common core, the model should abstain rather than hallucinate. Conversely, intersecting over too many environments risks a vacuous $a$, an instance of Nozick's tradeoff between breadth and depth. 
For normative questions, where no single objective core exists, \citet{sage} propose a complementary inference-time procedure---simulation-augmented generation (SAGE)---that queries simulated individuals from a prompt-appropriate population to preserve, rather than intersect away, the pluralistic variance across perspectives.

Which environments $\{e_i\}$ are admissible is, as discussed in Section~\ref{sec:admissible}, not given a priori. Today, this choice is mostly delegated to human operators, who encode their beliefs about relevant invariances through data curation, augmentation, and synthesis.\footnote{There are exceptions, such as the XRM algorithm~\citep{xrm} for automatically discovering environments from data.} The second step above opens a different possibility: the model itself proposes the environments---a technique sometimes referred to as \emph{reverse prompting}---and the resulting proposals $\{a_i\}$ can in turn reshape which environments are worth considering. This is Nozick's bootstrapping of facts and admissible transformations, carried out by the model itself, which becomes a participant in hypothesis testing and falsification. The invariant answer $a$, however, can only be as good as the array of environments $\{e_i\}$ that the model is able to generate,\footnote{Environments $e_i$ could also index different models, bringing us closer to the idea of model councils~\citep{perplexity2026modelcouncil}.} 
and remains bounded by its largely textual cognitive apparatus.
\footnote{Ludwig Wittgenstein (``The limits of my language mean the limits of my world'', \emph{Tractatus} 5.6) and the linguists Edward Sapir and Benjamin Lee Whorf have argued that the language a person speaks shapes and constrains their model of the world.}
Moreover, the contexts of a single model are not independent sources: they can expose framing artifacts, but not errors shared by its entire training corpus, which call for environments obtained from different models, retrieval, or experimentation with the world.

This procedure may succeed where IRM struggled (Section~\ref{sec:classifiers}) because of the output space. A classifier emits a label, which says little about whether it relied on invariant or spurious mechanisms. A language model emits a sequence $a_i$, which can expose the mechanism itself. In the notation of Section~\ref{sec:classifiers}, reasoning traces turn input-output pairs $(x, w(\phi(x)))$ into triplets $(x, \phi(x), w(\phi(x)))$ that annotate intermediate computations. Obtaining the answer ``It is a waterbird'' from a classifier is less informative than the LLM chain-of-thought ``I have no idea what type of bird this is, but I can see water in the background, so it is a waterbird.'' The latter reveals reliance on a spurious correlation, provided the chains-of-thought are faithful~\citep{turpin2023language, lanham2023measuring}. Cartwright makes the same move from predictions to explanations when she contrasts a derivation that merely fits the phenomena with a claim about their causes, illustrated with a plant whose leaves yellow because water in its tub rots its roots:
\begin{quote}
God tells you that Schr\"{o}dinger's equation provides a completely satisfactory derivation of the phenomenological law of radioactive decay. You have no doubt that the derivation is correct. But you still have no reason to believe in Schr\"{o}dinger's equation. On the other hand, if God tells you that the rotting of the [plant] roots is the cause of the yellowing of the leaves [\ldots] [t]hen you do have reason, conclusive reason, to believe that there is water in the tub. \citep[p.~93]{cartwright1983}
\end{quote}
In machine learning terms, \citet{kartikcot} show that, under suitable conditions, chain-of-thought annotations act as privileged information that reduces underspecification and promotes out-of-distribution generalization. Accordingly, when two proposals $a_i$ and $a_j$ disagree, the model, prompted with $\mathrm{inv}$, can inspect \emph{where} their reasoning diverges, rather than merely observing that it does. This is the structural realist lesson of Section~\ref{sec:philosophy} applied to language: invariance should be sought in the reasoning that produces an answer.

Finally, the optional fine-tuning step on $(q, a)$ suggests that invariance may offer a route to continual learning, viewed therefore as a problem of out-of-distribution generalization. Were the data distribution stationary, a model could keep learning from new data without suffering from catastrophic forgetting. In practice, user interactions induce a continually drifting distribution shaped, for instance, by world events. This suggests reframing continual learning as learning invariances across a different kind of environment, such as unrelated groups of users. These environments should not be time slices, since that would discard genuinely new patterns as non-invariant. If continual learning is the problem of deciding what new data is worth learning from, statistical invariance may serve as the guiding principle.

\section{Conclusion}

We have argued that likelihood, the foundational objective of modern language model training, is neither necessary nor sufficient for truth, and that \emph{invariance} across contexts is a better criterion, as Nozick, Wigner, and the structural realist tradition from Poincar\'{e} to Cassirer suggest. We have also sketched a test-time procedure that prompts a language model to propose multiple perspectives and extracts the common core of its answers.

The main counterpoint to these arguments is the ``scaling hypothesis'': given enough data, a model would observe every correlation under every environment, rendering explicit principles of invariance unnecessary. \citet{wigner1967} himself acknowledged this possibility:
\begin{quote}
If the universal law of nature should be discovered, the principles of invariance would lose their place in the hierarchy described before. [\ldots] If we had a complete description of all the events with which we shall ever come into contact, the correlations between these events, that is the laws of nature, would have a reduced significance, similar to that of the invariance principles when the universal law of nature is known. \citep[pp. 30--31]{wigner1967}
\end{quote}
However, the world keeps producing new environments, so deployment always happens partly out of distribution. Continual learning, perhaps via invariances, remains a requirement to build a more human-like artificial intelligence.

We should build systems that distinguish the ``probable'' from the ``true.'' As \citet[pp.~79--84]{nozick2001} observed, understanding objective facts is a stepwise process of mutually modifying knowledge of new facts and new admissible transformations. This process contrasts with local, gradient-based learning, because it allows radical theoretical shifts that, as \citet{gaukroger2012} argues, let us see ``the whole cosmos differently.''\footnote{\citet{churchland1979} offers a concrete exercise in shifting our perspective: using our position on Earth alongside two other planets on a clear night allows us to visually grasp our orientation relative to the Solar System plane.} The same bootstrapping process that drives scientific progress can guide the design of AI systems that look past the statistics of human discourse to the invariant structures underlying objectivity and truth---sometimes against all likelihood.

\clearpage
\newpage
\bibliographystyle{assets/plainnat}
\bibliography{references}

\end{document}